\def\year{2020}\relax
\documentclass[letterpaper]{article} 
\usepackage{aaai20}  
\usepackage{times}  
\usepackage{helvet} 
\usepackage{courier}  
\usepackage[hyphens]{url}  
\usepackage{graphicx} 
\usepackage{graphicx}  
\title{Dynamic Embedding on Textual Networks via a Gaussian Process }

\author{Pengyu Cheng, Yitong Li, Xinyuan Zhang,\smallskip \\ \Large \textbf{Liqun Chen,  David Carlson, Lawrence Carin} \smallskip \\ Duke University \smallskip \\pengyu.cheng@duke.edu}

\newcommand{\Emb}{\bm h}   
\newcommand{\tEmb}{\bm x}  
\newcommand{\sEmb}{\bm s}  

\newcommand{\tTxt}{\bm t}  

\newcommand{\citet }[1]{\citeauthor{#1} \shortcite{#1}}
\newcommand{\citep }{\cite}

\usepackage{amsmath,amsthm,amsfonts,amssymb}
\usepackage{bm}
\usepackage{sidecap}
\usepackage{mathrsfs}
\usepackage{multirow, multicol}
\usepackage{booktabs}

\newcommand{\calE}{\mathcal{E}}

\def\vtheta{{\bm{\theta}}}

\def\vg{{\bm{g}}}

\def\vs{{\bm{s}}}
\def\vt{{\bm{t}}}
\def\vu{{\bm{u}}}

\def\vx{{\bm{x}}}

\def\vz{{\bm{z}}}
\def\mA{{\bm{A}}}

\def\mK{{\bm{K}}}

\def\mS{{\bm{S}}}
\def\mT{{\bm{T}}}
\def\mU{{\bm{U}}}

\def\mX{{\bm{X}}}

\def\mZ{{\bm{Z}}}

\begin{document}

\maketitle

\begin{abstract}

Textual network embedding aims to learn low-dimensional representations of text-annotated nodes in a graph.
Prior work in this area has typically focused on fixed graph structures; however, real-world networks are often dynamic.
We address this challenge with a novel end-to-end node-embedding model, called Dynamic Embedding for Textual Networks with a Gaussian Process (DetGP). After training, DetGP can be applied efficiently to dynamic graphs without re-training or backpropagation.
The learned representation of each node is a combination of \emph{textual} and \emph{structural} embeddings. 
Because the structure is allowed to be dynamic, our method uses the Gaussian process to take advantage of its non-parametric properties.
To use both local and global graph structures, diffusion is used to model multiple hops between neighbors.
The relative importance of global versus local structure for the embeddings is learned automatically.
With the non-parametric nature of the Gaussian process, updating the embeddings for a changed graph structure requires only a forward pass through the learned model.
Considering link prediction and node classification, experiments demonstrate the empirical effectiveness of our method compared to baseline approaches.
We further show that DetGP can be straightforwardly and efficiently applied to dynamic textual networks.
\end{abstract}

\section{Introduction}

Learning latent representations for graph nodes has attracted considerable attention in machine learning, with applications in social networks~\citep{fan2019graph,perozzi2014deepwalk}, knowledge bases~\citep{trivedi2017know}, recommendation systems~\citep{ying2018graph}, and bioinformatics~\citep{zitnik2017predicting}. 
\emph{Textual networks} additionally contain rich semantic information, so text can be included with the graph structure to predict downstream tasks, such as link prediction~\citep{zhang2018diffusion}, node classification~\citep{kipf2016semi}, and graph generation~\citep{kipf2016variational}.
For instance, social networks have links between users, and typically each user has a profile (text). 
The goal of \emph{textual network embedding} is to learn node embeddings by jointly considering \emph{textual} and \emph{structural} information in the graph.

Most of the aforementioned textual network embedding methods focus on a fixed graph structure~\citep{tu2017cane,zhang2018diffusion,shen2018improved}. When new network nodes are added to the graph, these frameworks require that the whole model be re-trained to update the existing nodes and add representations for the new nodes, leading to high computational complexity.  However, networks are often dynamic; in social networks, users and relationships between users change over time ($e.g.$, new users, new friends, unfriending, $etc.$). It is impractical to update the full model whenever a new user is added. This paper seeks to address this challenge and learn an embedding method that adapts to a changed graph, without re-training.

Prior dynamic embedding methods usually focus on predicting how the graph structure changes over time, by training on multiple time steps ~\citep{zhou2018dynamic,seo2018structured,du2018dynamic}. In such a model, a dynamic network embedding is approximated by multiple steps of fixed network embeddings. In contrast, our method only needs to train on a single graph and it can quickly adapt to related graph structures.
Additionally, in prior work textual information is rarely included in dynamic graphs.
Two exceptions have looked at dynamic network embeddings with changing node attributes~\citep{li2017attributed,li2018streaming}. 
However, both require pre-trained node features, whereas we show that it is more powerful to learn the text encoder in a joint framework with the structural embedding.

We propose \textbf{D}ynamic \textbf{E}mbedding for \textbf{T}extual Networks with a \textbf{G}aussian \textbf{P}rocess (DetGP), a novel end-to-end model for performing unsupervised dynamic textual network embedding. DetGP learns \emph{textual} and \emph{structural} features jointly for both fixed and dynamic networks. The textual features indicate the intrinsic attributes of each node based on the text alone, while the structural features reveal the node relationships of the whole community. The structural features utilize both the textual features and the graph topology. This is achieved by smoothing the kernel function in a Gaussian process (GP) with a multi-hop graph transition matrix. This GP-based structure can handle newly added nodes or dynamic edges due to its non-parametric properties~\citep{rasmussen2003gaussian}. To facilitate fast computation, we learn inducing points to serve as landmarks in the GP \citep{yu2008gaussian}. Since the inducing points are fixed after training, computing new embeddings only requires calculating similarity to the inducing points, alleviating computational issues caused by changing graph structure.

To evaluate the proposed approach, the learned node embeddings are used for link prediction and node classification.
Performance on those downstream tasks demonstrates that learned embeddings capture relevant information. We also perform these tasks on dynamic textual networks and visualize the learned inducing points. Empirically, DetGP outperforms other models in downstream tasks, yielding efficient and accurate predictions on newly added nodes.

\section{Related Work}\label{sec:related_works}
\textbf{Textual Network Embedding}: Many graphs have rich text information for each node~\citep{yang2015network} (\textit{e.g.}, paper abstracts in citation networks, user profiles in social networks, product and costumer descriptions in online shopping, \textit{etc.}), which leads to the problem of \emph{textual network embedding}~\citep{tu2017cane}. Textual network embedding learns enhanced node representations by combining the textual embeddings and structural embeddings together. Textual  embeddings are learnt by encoding raw-text via text encoders. Structural are obtained based  on the connection information on graphs. Recent textual embedding methods~\citep{zhang2018diffusion,shen2018improved,tu2017cane,sun2016general} learn both textual and structural embeddings jointly in an end-to-end training process with neural networks. However, previous methods require all the network connection information before training, which makes them difficult to apply in dynamic network scenarios.

\textbf{Dynamic Network Embedding}: Graph structures are often \emph{dynamic} (\textit{e.g.}, paper citation increasing or social relationship changing overtime), but fixed network embedding algorithms require re-training when graphs change. This issue is addressed by dynamic network embeddings~\citep{trivedi2018representation}. Research on dynamic graphs has usually focused only on \emph{structural embeddings}~\citep{trivedi2018representation,du2018dynamic,guo2018spine} without considering rich side information associate with vertices. While these methods learn how to update the model as the graph changes, they are not real-time algorithms, instead using gradient back-propagation to update representations when new nodes are added or the graph structure changes~\citep{goyal2018dyngem,du2018dynamic}.

\textbf{Gaussian Process}: A Gaussian Process (GP) $f(\vx)$ is a collection of random  variables such that any finite subset of those variables are Gaussian distributed. More specifically, given finite set $\{ \vx_1, \vx_2, \dots, \vx_n \}$ with $n \in \mathbb{N}_+$, the corresponding signal $[f(\vx_1), f(\vx_2), \dots, f(\vx_n)]^\intercal \sim \mathcal{N}([m(\vx_1), m(\vx_2), \dots, m(\vx_n)]^\intercal, [k(\vx_i, \vx_j)]_{n \times n})$, where $m(\vx)$ is a mean function and $k(\cdot, \cdot)$ is a covariance kernel function. GPs are used widely in the Bayesian machine learning literature as priors on functions~\citep{titsias2009variational,rasmussen2003gaussian,titsias2010bayesian}. 
To learn the mapping $y = f(\vx)$ with training data $\{(\vx_i,y_i) \}_{i=1}^N$ and unlabeled testing data $\{ \vx'_j \}_{j=1}^M$, with Gaussian process as the prior, $[f(\vx_1), f(\vx_2) \dots, f(\vx_n), f(\vx'_1), \dots, f(\vx'_M)]$ follows a multivariate Gaussian distribution. Given observations $f(\vx_i) = y_i$, the conditional distribution for $[f(\vx'_1), \dots, f(\vx'_M)]$ can be easily obtained, which is also Gaussian distributed.
Recently, GPs have been used for node embedding learning on graphs~\citep{ma2018depthlgp,ng2018bayesian}. However, these works require pre-trained node features, leading to a two-step training process. In contrast, our proposed model is an end-to-end framework that can  jointly train the feature extractor (text encoder) with the GP parameters.

\section{Model}\label{sec:model}

We assume the input data are given as an undirected graph $\mathcal{G} = (\mathcal{V}, \mathcal{E})$, where $\mathcal{V}= \{v_n\}_{n=1}^N$ is the node set and $\calE = \{(v_n, v_{n'}) : v_n,v_{n'} \in \mathcal{V} \}$ is the edge set. Each node $v_n$ has an associated  $L_n$-length text sequence $\bm t_n = \{w_i \}_{i=1}^{L_n}$, where each $w_i$ is a natural language word. The adjacency matrix $\mA \in \{0,1\}^{N\times N}$ represents node relationships, where $A_{nn'} = 1$ if $(v_n, v_{n'}) \in \mathcal{E}$ and $A_{nn'} = 0$ otherwise. Our objective is to learn a low-dimensional embedding vector $\Emb_n$ for each node $v_n \in \mathcal{V}$ that captures both textual and structural features of the graph $\mathcal{G}$. 

\begin{figure*}[t]
    \centering
    \includegraphics[width=0.8\textwidth]{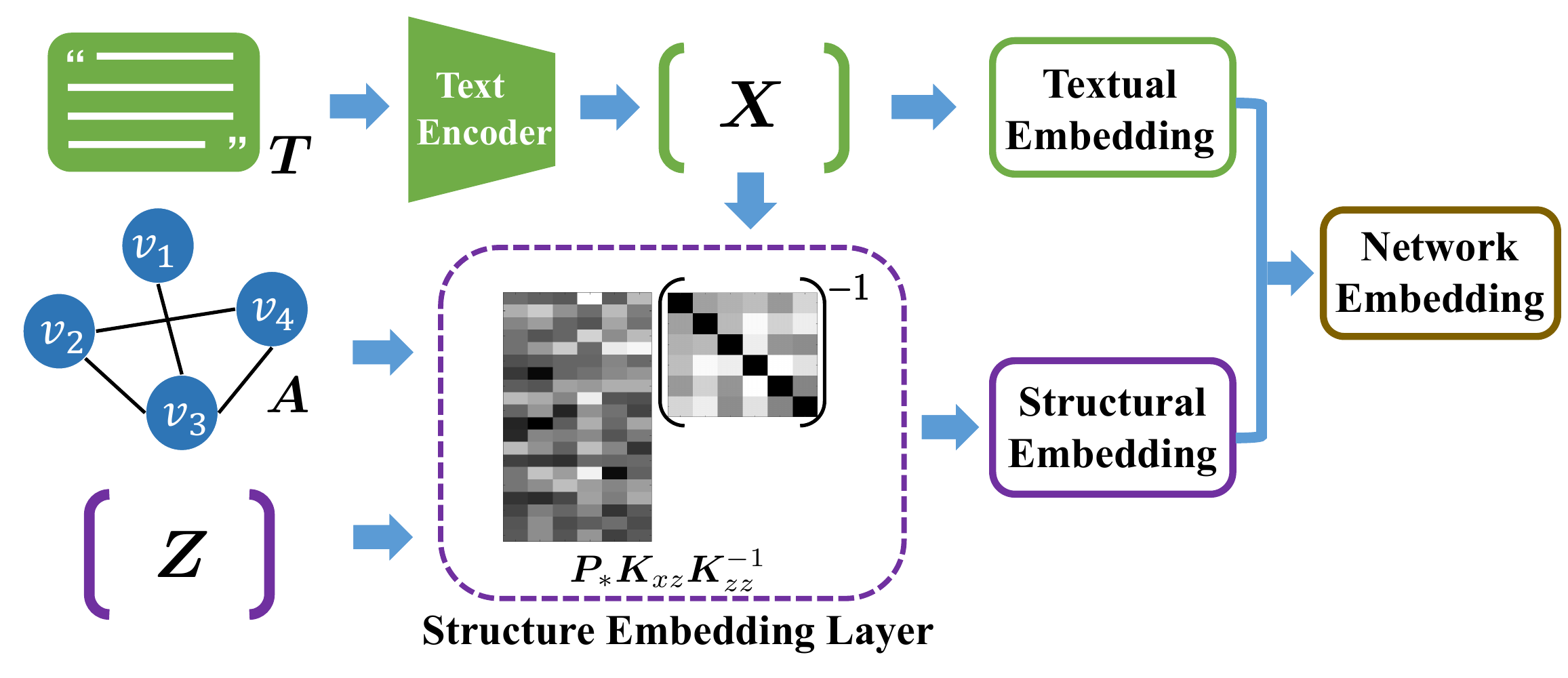}
 
    \caption{The input of the DetGP is the connection information $\mA$ of the network and textual side information $\mT = \{\vt_n\}_{n=1}^N$ for nodes. The model first encodes text $\mT$ to a low-dimension representation $\mX = \{\vx_n \}_{n=1}^N$, then infers the structural embeddings by $\mX$ and $\mA$ via a Gaussian process. Inducing points $\mZ = \{ \vz_m \}_{m=1}^M$ are used to reduce computational complexity. The output network embeddings combine the textual and structural embeddings. 
}
    \label{fig:framework}

\end{figure*}

Figure~\ref{fig:framework} gives the framework of the proposed model, DetGP. Text $\bm t_n$ is input to a text encoder $\bm g_{\vtheta }(\cdot)$ with parameters $\bm \theta$, described in Section \ref{subsec:text_encoder}. The output $\tEmb_n = \vg_\vtheta(\vt_n)$ is the \emph{textual embedding} of node $v_n$. This textual embedding is both part of the complete embedding $\Emb_n$ and an input into the structural embedding layer (dotted purple box) that is combined with the graph structure in a GP framework, discussed in Section \ref{subsec:embedding_layer}. In addition, multiple hops are modeled in this embedding layer to better reflect the graph architecture and use both local and global graph structure. The mathematical analysis of the structural embeddings is given in Section \ref{subsec:analysis_of_embedding_layer}. To scale up the model to large datasets, we adopt the idea of inducing points~\citep{titsias2009variational,titsias2010bayesian}, which serve as grid points in the model.  The output structural embeddings are denoted as $\sEmb_n$, which are combined to form the complete node embedding $\Emb_n=[\tEmb_n;\sEmb_n]$.

The model is trained by using the negative sampling loss~\citep{tu2017cane}, where neighbor nodes should be more similar than non-neighbor nodes. The learning procedure is fully described in Section \ref{subsec:algorithm_outline}. When the graph structure is updated, \emph{i.e.} new nodes $v_{new}$ with text $\tTxt_{new}$ are added or links are changed, the node embeddings are updated by a single forward-propagation step without relearning any model parameters. This property comes from the non-parametric nature of the GP-based structure, and it greatly increases efficiency for dynamic graphs. 

\subsection{Text Encoder}\label{subsec:text_encoder}

There are many existing text encoders~\citep{melamud2016context2vec,cer2018universal,lin2017structured}, often based on deep neural networks. However, using a deep neural network encoder can overfit on graphs because of the relatively small size of textual data \citep{zhang2018diffusion}. Therefore, various encoders are proposed to extract rich textual information specifically from graphs \citep{tu2017cane,zhang2018diffusion,shen2018improved}. In general, 
we aim to learn a text encoder $\vg_{\vtheta}$ with parameters $\vtheta$ that encodes semantic features $\vx_n = \vg_\vtheta(\vt_n)$. A simple and effective text encoder is the word embedding average (Wavg) $\vg_\vtheta(\vt)  = \frac{1}{L_n} \sum_{i=1}^{L_n} \bm \nu_{i}$, where $\bm \nu_{i}$ is the corresponding embedding of $w_{i}$ from the sequence $\bm t_n = \{w_i \}_{i=1}^{L_n}$.  This is implemented by a learnable look-up table.
~\citep{zhang2018diffusion} proposed a diffused word average encoder (DWavg) to leverage textual information over multiple hops on the network. Because DetGP focuses mainly on the structural embeddings, we do not focus on developing a new text encoder. Instead, we show that DetGP has compatibility with different text encoders, and our experiments use these two text encoders (Wavg and DWavg).

\subsection{Structural Embedding Layer}\label{subsec:embedding_layer}

The structural embedding layer transforms the encoded text feature $\tEmb$ to structural embedding $\sEmb$ using a GP in conjunction with the graph topology. Before introducing the GP, we introduce the multi-hop transition matrix $\bm P_*$ that will smooth the GP kernel.

\textbf{Multi-hop Transition Matrix}: In a graph, one node can be directly or indirectly connected to others by edges. Only considering direct connections is limiting. For example, in citation networks, cited papers should be closely related to a manuscript, so considering both a neighbor and its neighbors should add information. Following this intuition, we propose to use multiple hops, or multi-step graph random walks, to model both local and global structure.

Suppose $\bm P$ is the normalized transition matrix, $i.e.$ a normalized version of $\bm A$ where each row sums to one and $P_{nn'}$ represents the probability of a transition from node $n$ to node $n'$. If $\bm P$ represents the transition from a single hop, then higher orders of $\bm P$ will give multi-hop transition probabilities. Specifically, $\bm P^j$ is the $j$th power of $\bm P$, where $P^j_{nn'}$ gives the probability of transitioning from node $n$ to node $n'$ after $j$ random hops on the graph~\citep{abu2018watch}.
Different powers of $\bm P$ provide different levels of smoothing on the graph, and vary from using local to global structure. \emph{A priori} though, it is not clear what level of structure is most important for learning the embedding.  Therefore, they are combined in a learnable weighting scheme. 
Denoting the weights as $\bm \alpha$, this is
\begin{equation}\label{eq:P*}
   \textstyle \bm P_*=\sum_{j=0}^J \alpha_j \bm P^j , \quad s.t. \sum_{j=0}^J \alpha_j = 1,
\end{equation}
where $J$ is the maximum number of steps considered.
The constraint that $\sum_{j=0}^J \alpha_j = 1$ in \eqref{eq:P*} is implemented by a softmax function. Note that $\bm P^0 = \bm I_N$ is an identity matrix, that treats each node independently. In contrast, a large power of $\bm P$ would typically be very smooth after taking many hops. 
Therefore, $\bm \alpha$ can learn the importance of local ($\bm P^0$ or $\bm P^1$) and global (large powers of $\bm P$) graph structure for the node embeddings.  Equation~\eqref{eq:P*} can be viewed as a generalized form for DeepWalk~\citep{perozzi2014deepwalk} or Node2vec~\citep{grover2016node2vec}. In practice, learning the weights appears to be more robust than hand-engineering them~\citep{abu2018watch}.

\textbf{GP Structural Embedding Prior}: To describe the other key components of the structural embedding layer, we describe the GP approach. We define a latent function $f(\tEmb)$ over the textual embedding $\tEmb$ with a GP prior $f(\tEmb) \sim \mathcal{GP}\left( \bm 0, k(\tEmb_n, \tEmb_{n'}) \right)$.  Inspired by~\citep{ng2018bayesian}, instead of using this GP directly to determine the embedding, the learned graph diffusion is used on top of this Gaussian process. For finite samples, the combination of the graph diffusion and the GP yields a conditional structural embedding that can be expressed as a multivariate Gaussian distribution: 
\begin{small}
\begin{equation}\label{eq:original_gp}
    p([s_{1i},\dots,s_{Ni}]^\intercal| [\tEmb_1,\dots,\tEmb_N]) = \mathcal{N}\left(\bm 0, \bm P_* ^{\intercal} \bm K_{XX} \bm P_* \right) ,
\end{equation}
\end{small}
where $[\bm K_{XX}]_{nn'} = k(\tEmb_n, \tEmb_{n'})$ and $i$ is a index of our structure embedding feature.
Each dimension of the structural embedding follows this Gaussian distribution with the same covariance matrix $\bm K_{XX}$ smoothed by $\bm P^*$.
In practice, we use the first-degree polynomial kernel 
\begin{equation}\label{eq:kernel}
    k\left( \tEmb_n, \tEmb_{n'} \right) = \tEmb_n^{\intercal} \tEmb_{n'}+C , \ C>0
\end{equation}
because it outperforms others due to its numerical stability.
Moreover, the linear kernel in \eqref{eq:kernel} speeds up computation and increases model stability. 

\begin{figure}[t]
\centering
\includegraphics[width=0.6\columnwidth]{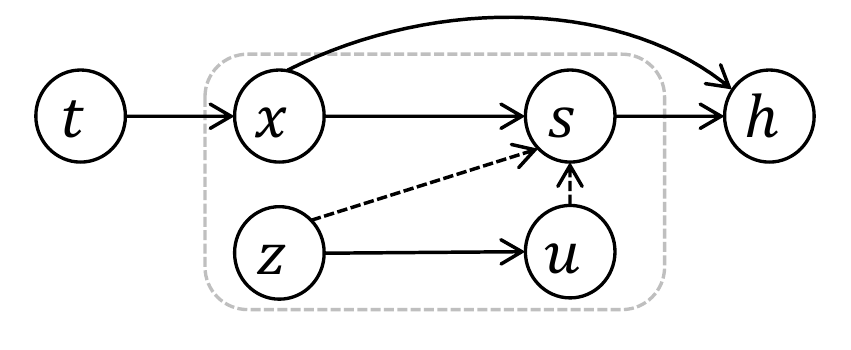}

\caption{\label{fig:notation_illustration}
Model flow chart. Inducing points are denoted as $\bm z$, with embedding $\bm u$. The final node embedding $\bm h$ is the concatenation of text embedding $\tEmb$ and structural embedding $\sEmb$.
}

\end{figure}

\textbf{Inducing Points}: GP models are known to suffer from computational complexity with large data size $N$. 
To scale up the model, we use the inducing points based on the Sparse pseudo-inputs Gaussian process (SPGP)~\citep{snelson2006sparse}. Let $\bm Z = [\bm z_1, \cdots, \bm z_M]^{\intercal}$ with $M<N$ denote inducing points (pseudo-textual embeddings) in the same space as the textual features. Assume $\bm U = [\bm u_1,\cdots,\bm u_M]^{\intercal}$ are corresponding the pseudo-structural embeddings of $\bm Z$, which is a function of $\bm z$ following the same GP function.
The structural and textual embeddings of real data samples are denoted as $\bm{S}=[\sEmb_1,\cdots, \sEmb_N]^{\intercal}$ and $\bm{X}=[\tEmb_1,\cdots, \tEmb_N]^{\intercal}$. Given the inducing points, the conditional distribution of our structural embeddings is
\begin{small}
\begin{align}\label{eq:e_x_conditional}
\textstyle
    &p(\bm S_i| \bm X, \bm Z, \bm U)  = \mathcal{N} \left( \bm \mu_{S_i|Z}, \bm \Sigma_{S|Z}\right), \\
    &\bm \mu_{S_i|Z}  = \bm P_*^{\intercal}  \bm K_{XZ}  \left(\bm K_{ZZ} + \sigma \bm I_{M}\right)^{-1} \bm U_i, \nonumber \\
    &\bm \Sigma_{S|Z} = \bm P_*^{\intercal} \bm K_{XX} \bm P_* - \bm P_*^{\intercal} \bm K_{XZ} \left( \bm K_{ZZ} + \sigma \bm I_M \right)^{-1} \bm K_{ZX} \bm P_*, \nonumber 
\end{align}
\end{small}
where $[\bm K_{XZ}]_{nm} = k(\tEmb_n, \vz_{m})$ and $[\bm K_{ZZ}]_{mm'} = k(\bm{z}_m, \bm{z}_{m'})$. The subscript $i$ indicates the $i$th column of a matrix ($\bm S_i$ is the concatenation of the $i$th element from all node structural embeddings). Each dimension of $\bm S$ has a multivariate Gaussian distribution with unique mean value $\bm \mu_{s_i|Z}$ but the same covariance $\bm \Sigma_{S|Z}$. Note that a small number $\sigma$ is added to the diagonal elements of the kernel $\mK_{ZZ}$ to enhance model stability. The relationships of different parameters are given in Figure~\ref{fig:notation_illustration}. The inducing points $\{\vz_m\}_{m=1}^M$ and text features $\{\vx_n \}_{n=1}^N$ share the same textual embedding space;  the pseudo-textual embeddings $\{ \vu_m\}_{m=1}^M$ and the structural embeddings $\{\vs_n\}_{n=1}^n$ share the same structural embedding space.

During the training, the textual embedding space is continuously changed, because the text encoder $\vg_\vtheta(\vt)$ is updated by the gradient descent algorithm iterative. Therefore, the inducing points in the textual embedding space should also be updated correspondingly. To obtain the optimal inducing points $\mZ$ with corresponding pseudo-structural embeddings $\mU$, we jointly train them using gradient descent with  weights in other layers. To back-propagate through the stochastic posterior distribution $p(\mS_i | \mX, \mZ, \mU)$ in equation~\eqref{eq:e_x_conditional}, we propose the mean approximation strategy.
We use the mean of $p(\mS_i | \mX, \mZ, \mU)$, $\hat{\mS} = \bm P_*^{\intercal} \bm K_{XZ} \left(\bm K_{ZZ} + \sigma \bm I_M\right)^{-1} \bm U$, as unbiased estimation of structural embeddings. Then gradients from $\mS$ can be easily propagated to $\mZ$ and $\mU$.  

\subsection{Analysis of the Structural Embedding}\label{subsec:analysis_of_embedding_layer}

We analyze the kernel function in \eqref{eq:original_gp} and~\eqref{eq:kernel} to show how the graph structure is used in the embedding layer. Denote $P^j_{nn'}$ in equation~\eqref{eq:P*} as the transition probability from node $n$ to node $n'$ in $j$ hops, then the correlation between node $n$ and $n'$ in \eqref{eq:original_gp} can be expanded as
\begin{small}
\begin{align}\label{eq:analyze_original}
\textstyle
     \text{cov}&(s_{ni}, s_{n'i}) = \alpha_0^2 k(\tEmb_{n}, \tEmb_{n'})+\alpha_0 \sum_{j=1}^J \alpha_j \sum_{r=1}^N P^j_{nr} k(\tEmb_n,  \tEmb_{r})\nonumber \\
     &+\alpha_0 \sum_{j=1}^J \alpha_j \sum_{r=1}^N P^j_{n'r} k(\tEmb_{n'},  \tEmb_{r})  \\
      &+ \sum_{j=1}^J \sum_{j'=1}^J  \sum_{\substack{r=1 \\ r\neq n, n}}^N \sum_{\substack{r'=1 \\ r' \neq n, n'}}^N \alpha_j \alpha_{j'} P^j_{nr} P^{j'}_{n'r'} k(\tEmb_{n'},  \tEmb_{r'}) \nonumber
\end{align}
\end{small}
The covariance is the same for all indices $i$.
The first term in \eqref{eq:analyze_original} measures the kernel function between $\tEmb_n$ and $\tEmb_{n'}$. The next two terms show the relationship between $\tEmb_n$ and the weighted multi-hop neighbors of $\tEmb_{n'}$ and vice versa. $\bm \alpha$ controls how much different hops are used. The last term is the pairwise-weighted higher order relationship between any two nodes in the graph, except $n$ and $n'$. 
The covariance structure uses the whole graph and learns how to balance local and global information. 
If node $n$ has no edges, then it will not be influenced by other nodes besides textual similarity. In contrast, a node with dense edge connections will be smoothed by its neighbors.

With the inducing points $[\vz_1,\vz_2,\dots,\vz_M]$, equation~\eqref{eq:analyze_original} can be modified as
 $  \text{cov}(s_{ni}, u_{mi}) = \alpha_0 k\left( \tEmb_{n}, \bm z_{m} \right) + \sum_{j=1}^J \alpha_j \sum_{n'=1}^N P^j_{nn'}k\left( \tEmb_{n'}, \bm z_{m} \right) .$
The covariance between node $v_n$ and the inducing points includes the local information $k\left( \tEmb_{n}, \bm z_{m} \right)$, as well as the smoothed effect from $\sum_{j=1}^J \alpha_j \sum_{n'=1}^N P^j_{nn'}k\left( \tEmb_{n'}, \bm z_{m} \right)$. This can also be viewed as feature smoothing over neighbors. Since inducing points do not contain links to other inducing points, there is no smoothing function for them. Each inducing point can be viewed as a node that already includes global graph information.

\subsection{Algorithm Outline}\label{subsec:algorithm_outline}

The structural embedding $\sEmb_n$ and the textual embedding $\tEmb_n$ are concatenated to form the final node embedding $\Emb_n = [\tEmb_n; \sEmb_n]$. If desired, $\Emb_n$ can then pass through several additional fully connected layers; for simplicity, we consider this the final embedding form. To learn the embeddings in the unsupervised framework, most existing works adopt the technique of negative sampling~\citep{zhang2018diffusion,zhang2018nscaching,tu2017cane}, that tries to maximize the conditional probability of one node's embedding given its one- and multi-hop neighbors, while maintaining a low conditional probability for non-neighbors. In the proposed framework, this loss is given for a single hop,
\begin{eqnarray}\label{eq:loss}
\textstyle
    \mathcal{L}=&-\frac{1}{|\mathcal{E}|}\sum_{(v_n,v_{n'}) \in \mathcal{E}} \log \left( \sigma(\bm h^{\intercal}_{n} \bm h_{n'}) \right)\nonumber \\ &- \frac{1}{N_s}\sum_{(v_n,v_{n'}) \not\in \mathcal{E}} \log \left[ 1-\sigma(\bm h^{\intercal}_{n} \bm h_{n'})  \right] .
\end{eqnarray}
$N_s$ is a weighting constant and can be set as  $N_s = \# \{(v_n,v_{n'}) \not\in \mathcal{E} \}$. In practice, using all nodes is infeasible, so a subset of neighbors and non-neighbors will be sampled. Equation~\eqref{eq:loss} maximizes the inner product among neighbors in the graph while minimizing the similarity among non-neighbors. 
Our model is trained end-to-end by taking gradients of loss $\mathcal{L}$  with respect to $\bm \theta$, $\bm Z$ and $\bm U$. The inducing points are initialized as the $k$-means centers of the encoded text features. Then, $\bm Z$ and the text encoder are trained jointly to minimize the loss function. Note that our model can also take a mini-batch of nodes as in GraphSage~\citep{hamilton2017inductive}. Adam~\citep{kingma2014adam} is used to optimize the parameters.

For a new node $v_{new}$ with text $\tTxt_{new}$, the transition matrix $\bm P_{new}$ is first updated, and the embeddings can be applied directly without additional back-propagation. Specifically, we first compute $\tEmb_{new} = \bm g_{\theta}(\tTxt_{new})$ from the text encoder. 
Then with $[\tEmb, \tEmb_{new}]$, the structural embedding of all nodes can be computed as $[\sEmb, \sEmb_{new}] = \sum_{j=0}^J \alpha_j \bm P^{j}_{new} \bm K_{x_{new}Z} (\bm K_{ZZ} + \sigma \bm I_M)^{-1} \bm U$. 
During this process, the structural embeddings of the original nodes also update due to the change in graph structure.

\section{Experiments}\label{sec:experiments}
\begin{table*}[hbt]  \footnotesize 
	\centering
	\scalebox{0.93}{

	\begin{tabular}{ccccccccccc}
        \toprule[1.2pt]
        \multicolumn{1}{c}{ }&  \multicolumn{5}{c}{Cora} & \multicolumn{5}{c}{HepTh}\\
        \hline
		\textbf{\%Training Edges} &  	\textbf{15\%} & \textbf{35\%} & 	\textbf{55\%} & \textbf{75\%} & \textbf{95\%} & \textbf{15\%} & \textbf{35\%} & \textbf{55\%} & \textbf{75\%} & \textbf{95\%}\\
		\midrule
		\textbf{MMB}~\citep{airoldi2008mixed}     & 54.7 & 59.5 & 64.9 & 71.1 &75.9 & 54.6 &57.3 & 66.2  &73.6 &80.3 \\
		\textbf{node2vec}~\citep{grover2016node2vec} & 55.9 & 66.1  & 78.7 & 85.9& 88.2 & 57.1 &69.9  &84.3 & 88.4 & 89.2 \\
		\textbf{LINE}~\citep{tang2015line}     & 55.0 & 66.4 & 77.6 & 85.6 & 89.3 & 53.7  &66.5  &78.5  &87.5 & 87.6\\ 
		\textbf{DeepWalk}~\citep{perozzi2014deepwalk} & 56.0 & 70.2 & 80.1 & 85.3 & 90.3 & 55.2 &70.0& 81.3  &87.6 & 88.0  \\ 
		\midrule 
		\textbf{TADW}~\citep{yang2015network} & 86.6 & 90.2 & 90.0 & 91.0& 92.7 & 87.0  &91.8& 91.1  &93.5 & 91.7 \\ 
		\textbf{CANE}~\citep{tu2017cane} & 86.8 &92.2 &94.6 &95.6 &97.7 & 90.0  &92.0  &94.2  &95.4 & 96.3\\
		\textbf{DMATE}~\citep{zhang2018diffusion}  & 91.3 & 93.7  & 96.0  & 97.4 & 98.8 & NA & NA & NA & NA & NA\\
		\textbf{WANE}~\citep{shen2018improved}  & 91.7 & 94.1 & 96.2 & \textbf{97.5} & \textbf{99.1} &92.3 & 95.7 & 97.5 & 97.7 & \textbf{98.7} \\
		\midrule
		\textbf{DetGP (Wavg) only Text } & 83.4& 89.1 & 89.9 & 90.9  & 92.3 &  86.5  & 89.6  & 90.2 & 91.5 & 92.6 \\
		\textbf{DetGP (Wavg) only Struct} & 85.4& 89.7 &91.0 & 92.7  & 94.1 &  89.7  & 92.1  & 93.5 & 94.8 & 95.1\\
		\textbf{DetGP (Wavg)} & 92.8& 94.8 &95.5 & 96.2  & 97.5 &  93.2  & 95.1  & 97.0 & 97.3 & 97.9\\
		  \midrule
		\textbf{DetGP (DWavg)} & \textbf{93.4} & \textbf{95.2} & \textbf{96.3}  & \textbf{97.5}& 98.8 & \textbf{94.3} &  \textbf{96.2}  &\textbf{97.7}  & \textbf{98.1} & {98.5}  \\
		\bottomrule[1.2pt]
	\end{tabular}}
	\caption{AUC scores for link prediction on the \emph{Cora} and \emph{HepTh} dataset. The top four models only have structural embedding, while the rest use text information.}
	\label{tab:link_predict_cora_hepth}
\end{table*}
\begin{table*}[htb]  \footnotesize 
	\centering
	\scalebox{0.93}{
	\begin{tabular}{ccccccccc}
        \toprule[1.2pt]
        \multicolumn{1}{c}{ }&  \multicolumn{4}{c}{Cora} & \multicolumn{4}{c}{DBLP}\\
        \hline
		\textbf{\%Training Nodes} &  	\textbf{10\%} & \textbf{30\%} & 	\textbf{50\%} & \textbf{70\%} & 	\textbf{10\%} & \textbf{30\%} & 	\textbf{50\%} & \textbf{70\%}\\
		\midrule
	
		 \textbf{LINE}\citep{tang2015line}  &  53.9 & 56.7 & 58.8 & 60.1  & 42.7 & 43.8 & 43.8 &  43.9\\
        \textbf{TADW}~\citep{yang2015network}  &  71.0 & 71.4 & 75.9 & 77.2  &67.6 & 68.9 & 69.2 &  69.5\\
        \textbf{CANE}~\citep{tu2017cane}  &  81.6 & 82.8 & 85.2 & 86.3 & 71.8 & 73.6 & 74.7 & 75.2\\
        \textbf{DMTE}~\citep{zhang2018diffusion}  &  81.8 & 83.9 & 86.3 & 87.9 &72.9 & 74.3 & 75.5 & 76.1\\
        \textbf{WANE}~\citep{shen2018improved}  &  81.9 & 83.9 & 86.4 & 88.1 & NA & NA &NA & NA \\
        \midrule
       \textbf{DetGP (Wavg) only Text  }  &  78.1 &  81.2    & 84.7 &  85.3 &71.4 &	73.3	&74.2	& 74.9   \\
       \textbf{DetGP (Wavg) only Struct}  &  70.9 &     79.7    & 81.5 &  82.3 &70.0&	71.4	&72.6	& 73.3   \\
        \textbf{DetGP (Wavg)}  &  80.5 &  85.4    & 86.7 &  88.5 &76.9 &	78.3	&79.1	& 79.3   \\
      \midrule
        \textbf{DetGP (DWavg)} & \textbf{83.1} & \textbf{87.2 }& \textbf{88.2}& \textbf{89.8} & \textbf{78.0} & \textbf{79.3} & \textbf{79.6}  & \textbf{79.8} \\
        \bottomrule[1.2pt]
	\end{tabular}}
	\caption{Test Macro-F1 scores for multi-label node classification on \emph{Cora} and  \emph{DBLP} dataset.}
	\label{tab:node_classification_cora_DBLP}
\end{table*}

To demonstrate the effectiveness of our DetGP embeddings, we conduct experiments first with static networks as in \citep{tu2017cane,zhang2018diffusion,shen2018improved}, and then on dynamic graphs. 
 In the standard setup, learning textual network embeddings requires that all nodes are available in the graph. In contrast, we demonstrate in Section~\ref{subsec:new_node_test} that our proposed model can infer embeddings efficiently after training on newly added nodes. We evaluate the graph embeddings on link prediction and node classification on the following real-world datasets:
%
\begin{itemize}
\item \textbf{Cora} is a paper citation network, with a total of 2,277 vertices and 5,214 edges in the graph, where only nodes with text are kept. Each node has a text abstract about machine learning and belongs to one of seven categories.
\item \textbf{DBLP} is a paper citation network with 60,744 nodes and 52,890 edges. Each node represents one paper in computer science in one of four categories: database, data mining, artificial intelligence, and computer vision.
\item \textbf{HepTh} (High Energy Physics Theory)~\citep{leskovec2007graph} is another paper citation network. The original dataset contains 9,877 nodes and 25,998 edges. We only keep nodes with associated text, so this is limited to 1,038 nodes and 1,990 edges.

\end{itemize}

For a fair comparison with previous work, we follow the setup in \citep{tu2017cane,zhang2018diffusion,shen2018improved}, where the embedding for each node has dimension 200, a concatenation of a 100-dimensional textual embedding and a 100-dimensional structural embedding. We evaluate our DetGP base on two text encoders: the word embedding average (Wavg) encoder and the diffused word embedding average (DWavg) encoder from \citet{zhang2018diffusion}, introduced in Section~\ref{subsec:text_encoder}. The maximum number of hops $J$ in $\bm P^*$ is set to $3$. 

\subsection{Link Prediction}\label{subsec:link_prediction}

The link prediction task seeks to infer if two nodes are connected, based on the learned embeddings. This standard task tests if the embedded node features contain graph connection information. For a given network, we randomly keep a certain percentage ($15\%$, $35\%$, $55\%$, $75\%$, $95\%$) of edges and learn embeddings. 
At test time, we calculate the inner product of pairwise node embedding. A large inner product value indicates a potential edge between two nodes. The AUC score~\citep{hanley1982meaning} is computed in this setting to evaluate the performance. 
The results are shown in Table~\ref{tab:link_predict_cora_hepth} on Cora and HepTh. Since the DBLP dataset only has 52,890 edges which is far too sparse compared with the node number 60,744, we do not evaluate the AUC score on it as a consequence of high variance from sampling edges.
The first four models only embed structural features, while the remaining alternatives use both textual and structural embeddings. We also provide the DetGP results of with only textual embeddings and only structure embeddings for ablation study. 

From Table~\ref{tab:link_predict_cora_hepth}, adding textual information in the embedding can improve the link prediction result by a large margin. Even using only textual embeddings, DetGP gains significant improvement compared with only structure-based methods, and achieves competitive performance compared with other text-based embedding methods. Using only structural information is slightly better than using only textual embeddings, since link prediction is a more structure-dependent task, which also indicates that DetGP~learns inducing points $\bm Z$ that can effectively represent the network structure.
Compared with other textual network embedding methods, DetGP has very competitive AUC scores, especially when only given a small percentage of edges. Noting that for our methods the text encoders come from the baselines Wavg and DWavg \citep{zhang2018diffusion}, the performance gain should come from the proposed structural embedding framework. 
\subsection{Node Classification}\label{subsec:node_classification}
Node classification tasks aim to predict the category labels of the nodes based on the network structure and the node attributes. For textual networks, 
node classification requires high-quality \emph{textual embeddings} because \emph{structural embeddings} alone do not accurately reflect node categories. Therefore, we only compare to methods designed for textual network embedding. 
After training converges, a linear SVM classifier is learned on the trained node embeddings and performance is estimated by a hold-out set. 
In Table~\ref{tab:node_classification_cora_DBLP}, we compare our methods (Wavg+DetGP, DWavg+DetGP) with recent textual network embedding methods under different proportions ($10\%$, $30\%$, $50\%$, $70\%$) of given nodes in the training set.
Following the setup in \citet{zhang2018diffusion}, the evaluation metric is Macro-F1 score~\citep{powers2011evaluation}.  We test on the Cora and DBLP datasets, which have group label information, where DetGP~yields the best performance under all situations. This demonstrates that the proposed model can learn both representative textual and structural embeddings. The ablation study results (only textual embeddings \textit{vs.} only structural embeddings) demonstrates that textual attributes are more important than edge connections in classification task.
To describe the effect of learning the weighting in the diffusion, for the experiment on Cora with $10\%$ nodes given for training, the learned weights in $\bm \alpha$ are $[0.58, 0.12, 0.24, 0.05]$. Thus, local and second order transition features are more important.

\subsection{Dynamic Network Embedding}\label{subsec:new_node_test}
\begin{table*}[tbh]  \footnotesize 
	\centering
		\scalebox{0.95}{
	\begin{tabular}{lcccccccc}
        \toprule[1.2pt]
                 &  \multicolumn{4}{c}{Cora} & \multicolumn{4}{c}{HepTh}\\
        \hline
		\textbf{\%Training Nodes} &  	\textbf{10\%} & \textbf{30\%} & 	\textbf{50\%} & \textbf{70\%} & 	\textbf{10\%} & \textbf{30\%} & 	\textbf{50\%} & \textbf{70\%}\\
        \midrule
        \textbf{Only Text (Wavg)}  &61.2 & 77.9& 87.9 & 90.3 & 68.3 &83.7 & 84.2&86.9\\
         \textbf{Neighbor-Aggregate (Max-Pooling)} & 54.6  & 69.1 &  78.7 & 87.3 & 59.6 & 78.3 & 79.9 & 80.7\\
       \textbf{Neighbor-Aggregate (Mean)} & 61.8  & 78.4 &  88.0 & 91.2 & 68.2 & 83.9 & 85.5 & 88.3\\
       \textbf{GraphSAGE (Max-Pooling)} &  62.1  &  78.6 &  88.6 & 92.4 &  68.4 & 85.8& 88.1 & 91.2\\
       \textbf{GraphSAGE (Mean)} &  62.2    & 79.1 & 88.9  & 92.6 &  69.1 & 85.9& 89.0 & 92.4\\
        \textbf{DetGP} & \textbf{62.9} &
        \textbf{81.1} & \textbf{90.9} & \textbf{93.0} &\textbf{ 70.7} &\textbf{86.6} &\textbf{90.7} & \textbf{93.3}\\
        \bottomrule[1.2pt]
	\end{tabular}}
	\caption{Test AUC scores for link prediction on \emph{Cora} and  \emph{HepTh} datasets.}
	\label{tab:new_node_link_prediction_dwavg}
\end{table*}
\begin{table*}[thb]  \footnotesize 
	\centering
		\scalebox{0.95}{
	\begin{tabular}{lcccccccc}
        \toprule[1.2pt]
   &  \multicolumn{4}{c}{Cora} & \multicolumn{4}{c}{DBLP}\\
        \hline
	 \textbf{\% Training Nodes}&  	\textbf{10\%} & \textbf{30\%} & 	\textbf{50\%} & \textbf{70\%} & 	\textbf{10\%} & \textbf{30\%} & 	\textbf{50\%} & \textbf{70\%}\\
        \midrule
        \textbf{Only Text (Wavg)}  &  60.2& 76.3 & 83.5 & 84.8 & 56.7& 67.9 & 70.4 & 73.5 \\
        
        \textbf{Neighbor-Aggregate (Max-Pooling)}  &55.8 & 70.2 & 78.4  & 80.5 & 51.8 & 60.5 & 68.3 & 70.6 \\
        \textbf{Neighbor-Aggregate (Mean)}  &60.1 & 77.2 & 84.1  & 85.0 & 56.8 & 68.2 & 71.3 & 74.7 \\
        
        \textbf{GraphSAGE (Max-Pooling)}  & 61.3 & 78.2& 85.1& 86.3&  58.9 & 69.1 & 72.4 & 74.9 \\ 
       
        \textbf{GraphSAGE (Mean)}  & 61.4 & 78.4& 85.5 &  \textbf{86.6}&  59.0 & 69.3 & 72.7  & 75.1\\   
          
        \textbf{DetGP} & \textbf{62.1} & \textbf{79.3} & \textbf{85.8} & \textbf{86.6} & \textbf{60.2} & \textbf{70.1} & \textbf{73.2} & \textbf{75.8}\\
        \bottomrule[1.2pt]
	\end{tabular}}
	\caption{Test Macro-F1 scores for multi-label node classification on \emph{Cora} and  \emph{DBLP} dataset.}
	\label{tab:new_node_classification_dwavg}
\end{table*}

One of the advantages of the proposed model is the ability to quickly estimate embeddings on newly added nodes. To test the effectiveness of our model, we split nodes into training and testing sets. The embedding model is learned only from training nodes with corresponding edges. To evaluate, we embed the testing nodes directly without updating the model parameters. In this section we mainly focus on the performance of dynamic structural embeddings. Therefore, the text encoder is fixed as word embedding average (Wavg) for simplicity.

Previous works \citep{tu2017cane,zhang2018diffusion,shen2018improved} on textual network embedding require the overall connection information to train the structural embedding, which cannot directly assign (without re-training) structural embeddings to a new coming node with connection information unknown during training. Therefore, the aforementioned methods cannot be applied to dynamic networks. To obtain comparable baselines to DetGP, we propose several strategies based on the idea of inductive graph representation learning (GraphSAGE)~\citep{hamilton2017inductive}, which generates embedding for new nodes by aggregating the neighbors' embeddings.  The two embedding assigning strategies are: (a) \textbf{Neighbor-Aggregate}: aggregating the \textit{structural embeddings} from the neighbors in the training set, as the structural embedding for the new node;  (b) \textbf{GraphSAGE}: aggregating the \textit{textual embeddings} from the neighbors, then passing through a fully-connected layer to get the new node's structural embedding. For neighborhood information aggregating, we use the mean aggregator and the max-pooling aggregator as mentioned in \citep{hamilton2017inductive}.

We evaluate the dynamic embeddings for test nodes on link prediction and node classification tasks. For both tasks, we split the nodes into training and testing sets with different proportions ($10$\%, $30$\%, $50$\%, $70$\%).
When embedding new testing nodes, only their textual attributes and connections with existing training nodes are provided. For the link prediction task, we predict the edges between testing nodes based on the inner product between their node embeddings; for node classification, an SVM classifier is trained based on embeddings of training nodes. When new nodes come, we first embed the nodes using the trained model and then use the previously learned SVM to predict the label.

The results of link prediction and node classification are given in Tables~\ref{tab:new_node_link_prediction_dwavg} and~\ref{tab:new_node_classification_dwavg}, respectively. In both tasks, the Neighbor-Aggregate strategy with mean aggregator shows slight improvement to the baseline with only a text encoder. However, it does not work well with the max-pooling aggregator, implying that the unsupervised max-pooling on pre-trained neighbor structural embeddings cannot learn a good representation. The GraphSAGE strategies (with both mean and pooling aggregator) show notable improvements compared with Wavg and Neighbor-Aggregate. Unlike the unsupervised pooling, the GraphSAGE pooling aggregator is trained with a fully-connected layer on top, which shows comparable result to the mean aggregator. The proposed DetGP significantly outperforms other baselines, especially when the proportion of training set is small. A reasonable explanation is, when the training set is small, new nodes will have few connections with the training nodes, which causes high variance in the results of aggregating neighborhood embeddings. However, instead of aggregating, our DetGP infers the structural embedding via a Gaussian process with pre-learned inducing points, which is more robust than the information passed by neighbor nodes.

\subsection{Inducing Points}\label{subsec:inducing_points}

\begin{figure}[htb]
\centering
    \centering
    \includegraphics[width=0.5\columnwidth]{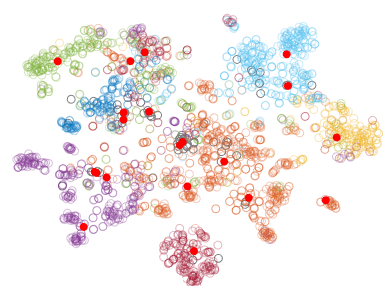}
    \caption{t-SNE visualization of learned network structural embeddings of Cora. Unfilled circles are individual nodes and inducing points are denoted as red dots.}
    \label{fig:tsne}
\end{figure}

Figure~\ref{fig:tsne} gives the t-SNE visualization of the learned DetGP structural embeddings on the Cora citation dataset. The model is learned using all edges and all of the nodes with their textual information.  We set the number of inducing points to $M=20$. To avoid the computational instability caused by the inverse matrix $\mK_{ZZ}^{-1}$, we update inducing points with a smaller learning rate, which is set to one-tenth of the learning rate for the text encoder. The inducing points $\bm z$ are visualized as red filled circles in Figure~\ref{fig:tsne}. Textual embeddings are plotted with $7$ different colors, representing the $7$ node classes. Note that the inducing points $\bm z$ fully cover the space of the categories, implying that the learned inducing points meaningfully cover the distribution of the textual embeddings.

\section{Conclusions}\label{sec:conclusion}

We propose a novel textual network embedding framework that learns representative node embeddings for static textual network, and also effectively adapts to dynamic graph structures. This is achieved by introducing a GP network structural embedding layer, that first maps each node to the inducing points, and then embeds them by taking advantage of the non-parametric representation. We also consider multiple hops to weight local and global graph structures. The graph structure is injected in the kernel matrix, where the kernel between two nodes use the whole graph information based on multiple hops. Our final embedding contains both \emph{structural} and \emph{textual} information. Empirical results demonstrate the effectiveness of the proposed algorithm.\\

\noindent{\bf Acknowledgements}: The research reported here was supported in part by DARPA, DOE, NIH, NSF and ONR.

\bibliographystyle{aaai}
{\small
\bibliography{ref}

\begin{thebibliography}{}

\bibitem[\protect\citeauthoryear{Abu-El-Haija \bgroup et al\mbox.\egroup
  }{2018}]{abu2018watch}
Abu-El-Haija, S.; Perozzi, B.; Al-Rfou, R.; and Alemi, A.~A.
\newblock 2018.
\newblock Watch your step: Learning node embeddings via graph attention.
\newblock In {\em NeurIPS},  9180--9190.

\bibitem[\protect\citeauthoryear{Airoldi \bgroup et al\mbox.\egroup
  }{2008}]{airoldi2008mixed}
Airoldi, E.~M.; Blei, D.~M.; Fienberg, S.~E.; and Xing, E.~P.
\newblock 2008.
\newblock Mixed membership stochastic blockmodels.
\newblock {\em JMLR}  1981--2014.

\bibitem[\protect\citeauthoryear{Cer \bgroup et al\mbox.\egroup
  }{2018}]{cer2018universal}
Cer, D.; Yang, Y.; Kong, S.-y.; Hua, N.; Limtiaco, N.; John, R.~S.; Constant,
  N.; Guajardo-Cespedes, M.; Yuan, S.; Tar, C.; et~al.
\newblock 2018.
\newblock Universal sentence encoder.
\newblock {\em arXiv preprint arXiv:1803.11175}.

\bibitem[\protect\citeauthoryear{Du \bgroup et al\mbox.\egroup
  }{2018}]{du2018dynamic}
Du, L.; Wang, Y.; Song, G.; Lu, Z.; and Wang, J.
\newblock 2018.
\newblock Dynamic network embedding: An extended approach for skip-gram based
  network embedding.
\newblock In {\em IJCAI},  2086--2092.

\bibitem[\protect\citeauthoryear{Fan \bgroup et al\mbox.\egroup
  }{2019}]{fan2019graph}
Fan, W.; Ma, Y.; Li, Q.; He, Y.; Zhao, E.; Tang, J.; and Yin, D.
\newblock 2019.
\newblock Graph neural networks for social recommendation.
\newblock {\em arXiv preprint arXiv:1902.07243}.

\bibitem[\protect\citeauthoryear{Goyal \bgroup et al\mbox.\egroup
  }{2018}]{goyal2018dyngem}
Goyal, P.; Kamra, N.; He, X.; and Liu, Y.
\newblock 2018.
\newblock Dyngem: Deep embedding method for dynamic graphs.
\newblock {\em arXiv preprint arXiv:1805.11273}.

\bibitem[\protect\citeauthoryear{Grover and
  Leskovec}{2016}]{grover2016node2vec}
Grover, A., and Leskovec, J.
\newblock 2016.
\newblock node2vec: Scalable feature learning for networks.
\newblock In {\em SIGKDD},  855--864.
\newblock ACM.

\bibitem[\protect\citeauthoryear{Guo, Xu, and Chen}{2018}]{guo2018spine}
Guo, J.; Xu, L.; and Chen, E.
\newblock 2018.
\newblock Spine: structural identity preserved inductive network embedding.
\newblock {\em arXiv preprint arXiv:1802.03984}.

\bibitem[\protect\citeauthoryear{Hamilton, Ying, and
  Leskovec}{2017}]{hamilton2017inductive}
Hamilton, W.; Ying, Z.; and Leskovec, J.
\newblock 2017.
\newblock Inductive representation learning on large graphs.
\newblock In {\em Advances in Neural Information Processing Systems},
  1024--1034.

\bibitem[\protect\citeauthoryear{Hanley and McNeil}{1982}]{hanley1982meaning}
Hanley, J.~A., and McNeil, B.~J.
\newblock 1982.
\newblock The meaning and use of the area under a receiver operating
  characteristic (roc) curve.
\newblock {\em Radiology} 143(1):29--36.

\bibitem[\protect\citeauthoryear{Kingma and Ba}{2015}]{kingma2014adam}
Kingma, D.~P., and Ba, J.
\newblock 2015.
\newblock Adam: A method for stochastic optimization.
\newblock {\em ICLR}.

\bibitem[\protect\citeauthoryear{Kipf and Welling}{2016}]{kipf2016variational}
Kipf, T.~N., and Welling, M.
\newblock 2016.
\newblock Variational graph auto-encoders.
\newblock {\em arXiv preprint arXiv:1611.07308}.

\bibitem[\protect\citeauthoryear{Kipf and Welling}{2017}]{kipf2016semi}
Kipf, T.~N., and Welling, M.
\newblock 2017.
\newblock Semi-supervised classification with graph convolutional networks.
\newblock {\em ICLR}.

\bibitem[\protect\citeauthoryear{Leskovec, Kleinberg, and
  Faloutsos}{2007}]{leskovec2007graph}
Leskovec, J.; Kleinberg, J.; and Faloutsos, C.
\newblock 2007.
\newblock Graph evolution: Densification and shrinking diameters.
\newblock {\em TKDD}.

\bibitem[\protect\citeauthoryear{Li \bgroup et al\mbox.\egroup
  }{2017}]{li2017attributed}
Li, J.; Dani, H.; Hu, X.; Tang, J.; Chang, Y.; and Liu, H.
\newblock 2017.
\newblock Attributed network embedding for learning in a dynamic environment.
\newblock In {\em Proceedings of the 2017 ACM on Conference on Information and
  Knowledge Management},  387--396.
\newblock ACM.

\bibitem[\protect\citeauthoryear{Li \bgroup et al\mbox.\egroup
  }{2018}]{li2018streaming}
Li, J.; Cheng, K.; Wu, L.; and Liu, H.
\newblock 2018.
\newblock Streaming link prediction on dynamic attributed networks.
\newblock In {\em Proceedings of the Eleventh ACM International Conference on
  Web Search and Data Mining},  369--377.
\newblock ACM.

\bibitem[\protect\citeauthoryear{Lin \bgroup et al\mbox.\egroup
  }{2017}]{lin2017structured}
Lin, Z.; Feng, M.; Santos, C. N.~d.; Yu, M.; Xiang, B.; Zhou, B.; and Bengio,
  Y.
\newblock 2017.
\newblock A structured self-attentive sentence embedding.
\newblock {\em ICLR}.

\bibitem[\protect\citeauthoryear{Ma, Cui, and Zhu}{2018}]{ma2018depthlgp}
Ma, J.; Cui, P.; and Zhu, W.
\newblock 2018.
\newblock Depthlgp: learning embeddings of out-of-sample nodes in dynamic
  networks.
\newblock In {\em AAAI}.

\bibitem[\protect\citeauthoryear{Melamud, Goldberger, and
  Dagan}{2016}]{melamud2016context2vec}
Melamud, O.; Goldberger, J.; and Dagan, I.
\newblock 2016.
\newblock context2vec: Learning generic context embedding with bidirectional
  lstm.
\newblock In {\em SIGNLL},  51--61.

\bibitem[\protect\citeauthoryear{Ng, Colombo, and Silva}{2018}]{ng2018bayesian}
Ng, Y.~C.; Colombo, N.; and Silva, R.
\newblock 2018.
\newblock Bayesian semi-supervised learning with graph gaussian processes.
\newblock In {\em NeurIPS}.

\bibitem[\protect\citeauthoryear{Perozzi, Al-Rfou, and
  Skiena}{2014}]{perozzi2014deepwalk}
Perozzi, B.; Al-Rfou, R.; and Skiena, S.
\newblock 2014.
\newblock Deepwalk: Online learning of social representations.
\newblock In {\em SIGKDD},  701--710.
\newblock ACM.

\bibitem[\protect\citeauthoryear{Powers}{2011}]{powers2011evaluation}
Powers, D.~M.
\newblock 2011.
\newblock Evaluation: from precision, recall and f-measure to roc,
  informedness, markedness and correlation.
\newblock {\em Bioinfo Publications}.

\bibitem[\protect\citeauthoryear{Rasmussen}{2006}]{rasmussen2003gaussian}
Rasmussen, C.~E.
\newblock 2006.
\newblock Gaussian processes for machine learning.
\newblock In {\em Springer}.

\bibitem[\protect\citeauthoryear{Seo \bgroup et al\mbox.\egroup
  }{2018}]{seo2018structured}
Seo, Y.; Defferrard, M.; Vandergheynst, P.; and Bresson, X.
\newblock 2018.
\newblock Structured sequence modeling with graph convolutional recurrent
  networks.
\newblock In {\em International Conference on Neural Information Processing},
  362--373.
\newblock Springer.

\bibitem[\protect\citeauthoryear{Shen \bgroup et al\mbox.\egroup
  }{2019}]{shen2018improved}
Shen, D.; Zhang, X.; Henao, R.; and Carin, L.
\newblock 2019.
\newblock Improved semantic-aware network embedding with fine-grained word
  alignment.
\newblock {\em EMNLP}.

\bibitem[\protect\citeauthoryear{Snelson and
  Ghahramani}{2006}]{snelson2006sparse}
Snelson, E., and Ghahramani, Z.
\newblock 2006.
\newblock Sparse gaussian processes using pseudo-inputs.
\newblock In {\em Advances in neural information processing systems},
  1257--1264.

\bibitem[\protect\citeauthoryear{Sun \bgroup et al\mbox.\egroup
  }{2016}]{sun2016general}
Sun, X.; Guo, J.; Ding, X.; and Liu, T.
\newblock 2016.
\newblock A general framework for content-enhanced network representation
  learning.
\newblock {\em arXiv preprint arXiv:1610.02906}.

\bibitem[\protect\citeauthoryear{Tang \bgroup et al\mbox.\egroup
  }{2015}]{tang2015line}
Tang, J.; Qu, M.; Wang, M.; Zhang, M.; Yan, J.; and Mei, Q.
\newblock 2015.
\newblock Line: Large-scale information network embedding.
\newblock In {\em Proceedings of the 24th international conference on world
  wide web},  1067--1077.
\newblock International World Wide Web Conferences Steering Committee.

\bibitem[\protect\citeauthoryear{Titsias and
  Lawrence}{2010}]{titsias2010bayesian}
Titsias, M., and Lawrence, N.~D.
\newblock 2010.
\newblock Bayesian gaussian process latent variable model.
\newblock In {\em AISTATS},  844--851.

\bibitem[\protect\citeauthoryear{Titsias}{2009}]{titsias2009variational}
Titsias, M.
\newblock 2009.
\newblock Variational learning of inducing variables in sparse gaussian
  processes.
\newblock In {\em AISTATS},  567--574.

\bibitem[\protect\citeauthoryear{Trivedi \bgroup et al\mbox.\egroup
  }{2017}]{trivedi2017know}
Trivedi, R.; Dai, H.; Wang, Y.; and Song, L.
\newblock 2017.
\newblock Know-evolve: Deep temporal reasoning for dynamic knowledge graphs.
\newblock In {\em Proceedings of the 34th International Conference on Machine
  Learning-Volume 70},  3462--3471.
\newblock JMLR. org.

\bibitem[\protect\citeauthoryear{Trivedi \bgroup et al\mbox.\egroup
  }{2018}]{trivedi2018representation}
Trivedi, R.; Farajtabar, M.; Biswal, P.; and Zha, H.
\newblock 2018.
\newblock Representation learning over dynamic graphs.
\newblock {\em arXiv preprint arXiv:1803.04051}.

\bibitem[\protect\citeauthoryear{Tu \bgroup et al\mbox.\egroup
  }{2017}]{tu2017cane}
Tu, C.; Liu, H.; Liu, Z.; and Sun, M.
\newblock 2017.
\newblock Cane: Context-aware network embedding for relation modeling.
\newblock In {\em ACL}.

\bibitem[\protect\citeauthoryear{Yang \bgroup et al\mbox.\egroup
  }{2015}]{yang2015network}
Yang, C.; Liu, Z.; Zhao, D.; Sun, M.; and Chang, E.
\newblock 2015.
\newblock Network representation learning with rich text information.
\newblock In {\em Twenty-Fourth International Joint Conference on Artificial
  Intelligence}.

\bibitem[\protect\citeauthoryear{Ying \bgroup et al\mbox.\egroup
  }{2018}]{ying2018graph}
Ying, R.; He, R.; Chen, K.; Eksombatchai, P.; Hamilton, W.~L.; and Leskovec, J.
\newblock 2018.
\newblock Graph convolutional neural networks for web-scale recommender
  systems.
\newblock In {\em SIGKDD},  974--983.
\newblock ACM.

\bibitem[\protect\citeauthoryear{Yu and Chu}{2008}]{yu2008gaussian}
Yu, K., and Chu, W.
\newblock 2008.
\newblock Gaussian process models for link analysis and transfer learning.
\newblock In {\em NIPS},  1657--1664.

\bibitem[\protect\citeauthoryear{Zhang \bgroup et al\mbox.\egroup
  }{2018a}]{zhang2018diffusion}
Zhang, X.; Li, Y.; Shen, D.; and Carin, L.
\newblock 2018a.
\newblock Diffusion maps for textual network embedding.
\newblock In {\em NeurIPS}.

\bibitem[\protect\citeauthoryear{Zhang \bgroup et al\mbox.\egroup
  }{2018b}]{zhang2018nscaching}
Zhang, Y.; Yao, Q.; Shao, Y.; and Chen, L.
\newblock 2018b.
\newblock Nscaching: Simple and efficient negative sampling for knowledge graph
  embedding.
\newblock {\em arXiv preprint arXiv:1812.06410}.

\bibitem[\protect\citeauthoryear{Zhou \bgroup et al\mbox.\egroup
  }{2018}]{zhou2018dynamic}
Zhou, L.; Yang, Y.; Ren, X.; Wu, F.; and Zhuang, Y.
\newblock 2018.
\newblock Dynamic network embedding by modeling triadic closure process.
\newblock In {\em Thirty-Second AAAI Conference on Artificial Intelligence}.

\bibitem[\protect\citeauthoryear{Zitnik and
  Leskovec}{2017}]{zitnik2017predicting}
Zitnik, M., and Leskovec, J.
\newblock 2017.
\newblock Predicting multicellular function through multi-layer tissue
  networks.
\newblock {\em Bioinformatics} 33(14):i190--i198.

\end{thebibliography}
}

\end{document}